\documentclass{article}

\PassOptionsToPackage{numbers, compress}{natbib}
\usepackage[final]{neurips_2023}
\usepackage[utf8]{inputenc} % allow utf-8 input
\usepackage[T1]{fontenc}    % use 8-bit T1 fonts
\usepackage{hyperref}       % hyperlinks
\usepackage{url}            % simple URL typesetting
\usepackage{booktabs}       % professional-quality tables
\usepackage{amsfonts}       % blackboard math symbols
\usepackage{nicefrac}       % compact symbols for 1/2, etc.
\usepackage{microtype}      % microtypography
\usepackage{xcolor}         % colors

\usepackage{nameref}
\usepackage{import}

\usepackage[numbers]{natbib}
\usepackage{float}
\usepackage{graphicx}
\usepackage{subcaption}

\usepackage{amssymb}
\usepackage{pifont}
\usepackage{tikz}

\usepackage{color}
\usepackage{amsmath}
\usepackage{amssymb}
\usepackage{amsthm}
\usepackage{mathtools}
\usepackage[mathscr]{eucal}% \mathscr{ABCDEFGHIJKLMNOPQRSTUVWXYZ}
\usepackage{bm}% bold math
\usepackage{bbm}
\usepackage{relsize}
\usepackage{graphics}
\usepackage{wrapfig}
\usepackage{multirow}
\usepackage{enumitem} 

\usepackage{ml_defs}    % using ml_defs package from the institute

\usepackage{array}
\usepackage{multicol}

\usepackage{pgf}
\usepackage{xinttools}%,pgfkeys,pgfmath}
\usepackage{tikz}
\usetikzlibrary{arrows.meta}
\usepackage{textcomp}
\usetikzlibrary{calc,shapes,arrows,positioning,automata,trees}
\usepackage{placeins} % Provides \FloatBarrier
\usepackage{tabularx}
\usepackage{graphicx}
\usepackage{subcaption}
\usepackage{xcolor}
\usepackage{listings}
\usepackage{xargs,lipsum,changepage,ifthen} % package `caption` could not be recognized

\usepackage[toc,page]{appendix}
\usepackage{selectp}
\newtheorem*{theorem*}{Theorem}
\newtheorem*{definition*}{Definition}

\renewcommand{\leq}{\leqslant}

\usepackage{pdflscape} % For landscape tables

\newcolumntype{R}[1]{>{\raggedright\arraybackslash}p{#1}}
\newcolumntype{C}[1]{>{\centering\arraybackslash}p{#1}}
\newcolumntype{L}[1]{>{\raggedleft\arraybackslash}p{#1}}

\usepackage{bigdelim}
\usepackage{colortbl} % colored cells
\definecolor{mColor1}{rgb}{0.95,0.95,0.95}
\definecolor{calRed}{rgb}{0.85, 0.51, 0.51} % 0.8, 0.51, 0.51
\definecolor{calGreen}{rgb}{0.7, 0.9, 0.65} % 0.7, 0.78, 0.65

\newcommand{\stoptocwriting}{%
  \addtocontents{toc}{\protect\setcounter{tocdepth}{-5}}}
\newcommand{\resumetocwriting}{%
  \addtocontents{toc}{\protect\setcounter{tocdepth}{\arabic{tocdepth}}}}

 \hypersetup{
   colorlinks=true,
   linkcolor=blue,
   urlcolor=magenta,
   pdfborder=0 0 0,
   pdftitle=Pretraining for Learning Abstractions,
   pdfsubject= {RUDDER, reinforcement learning, reward redistribution, return decomposition, delayed reward, sparse reward, episodic reward},
   pdfkeywords={RUDDER, reinforcement learning, reward redistribution, return decomposition, delayed reward, sparse reward, episodic reward, minecraft},
   pdfauthor= {},
   pdfstartview=FitH
}

\def\mathcolor#1#{\mathcoloraux{#1}}
\newcommand*{\mathcoloraux}[3]{%
  \protect\leavevmode
  \begingroup
    \color#1{#2}#3%
  \endgroup
}

\makeatletter
\newcommand{\setword}[2]{%
  \phantomsection
  #1\def\@currentlabel{\unexpanded{#1}}\label{#2}%
}
\makeatother

\title{Contrastive Abstraction for Reinforcement Learning}

\author{%
  Vihang Patil\thanks{Corresponding Author: patil@ml.jku.at} \quad Markus Hofmarcher \quad Elisabeth Rumetshofer \quad \textbf{Sepp Hochreiter} \\
    LIT AI Lab, Ellis Unit Linz and Institute for Machine Learning\\
    Johannes Kepler University Linz \\
    \texttt{\{patil, hofmarch, rumetshofer, hochreit\}@ml.jku.at}\\
}

\begin{document}

\maketitle

\begin{abstract}

Learning agents with reinforcement learning is difficult 
when dealing with long trajectories that involve a large number of states. 
To address these learning problems effectively, 
the number of states can be reduced by abstract representations that cluster states.
In principle, deep reinforcement learning can find abstract states, 
but end-to-end learning is unstable.
We propose \textit{contrastive abstraction learning} to find abstract states, 
where we assume that successive states in a trajectory 
belong to the same abstract state.
Such abstract states may be basic locations, achieved subgoals, 
inventory, or health conditions.
\textit{Contrastive abstraction learning} first constructs clusters of state representations
by contrastive learning and then applies modern Hopfield networks to determine the abstract states.
The first phase of \textit{contrastive abstraction learning} is self-supervised learning, 
where contrastive learning
forces states with sequential proximity
to have similar representations.
The second phase uses modern Hopfield networks to map 
similar state representations to the same fixed point, i.e.\ to an abstract state. 
The level of abstraction can be adjusted by determining the number of fixed points 
of the modern Hopfield network.
Furthermore, \textit{contrastive abstraction learning} does not require rewards 
and facilitates efficient reinforcement learning for a wide range of downstream tasks. 
Our experiments demonstrate the effectiveness of \textit{contrastive abstraction learning} for reinforcement learning.
\end{abstract}

\stoptocwriting
\section{Introduction}

Key in reinforcement learning (RL) is to learn 
proper representation of the environment.
If the state space 
is small and trajectories are short, 
RL can efficiently construct plans and solve tasks.
In particular, for Markov decision processes (MDPs) \cite{Puterman:05}
with few states and short trajectories,
an agent can readily learn world models, value functions, and policies.
Therefore, the main goal in representation learning for RL is to find
clusters of similar states, that form abstract states. 
If we know how to transit from one abstract state to another one,
then state abstraction transforms a complex problem into a simpler problem,
e.g.\ an MDP with many states into an MDP with few states.
Deep reinforcement learning gave us the hope that it
automatically identifies abstract states,
however, end-to-end learning of representations is not stable \cite{Laskin:20,Kostrikov:20, Steinparz:22}. 
Therefore, recent work revisited auxiliary losses and data augmentation to obtain
good representations \cite{Eysenbach:22}.
Also, learning proper state abstractions is computationally expensive, since
a proper clustering of states must be identified in large set of possible clusterings.
However, if state abstractions are learned reward free they can be used for many RL tasks, 
thus amortizing the cost of finding these state clusterings. 

A promising direction to learn proper RL representations without using rewards is
self-supervised learning, where  a learning system is trained to
capture the mutual dependencies between its inputs \cite{LeCun:22}. 
The data stream and not the model designer should determine 
good representations, which will emerge under appropriate training schemes \cite{Wang:22}. We leverage such reward free learning to obtain abstract states to make learning efficient in downstream tasks.

\begin{figure*}[t]
    \centering
    \includegraphics[width=\linewidth]{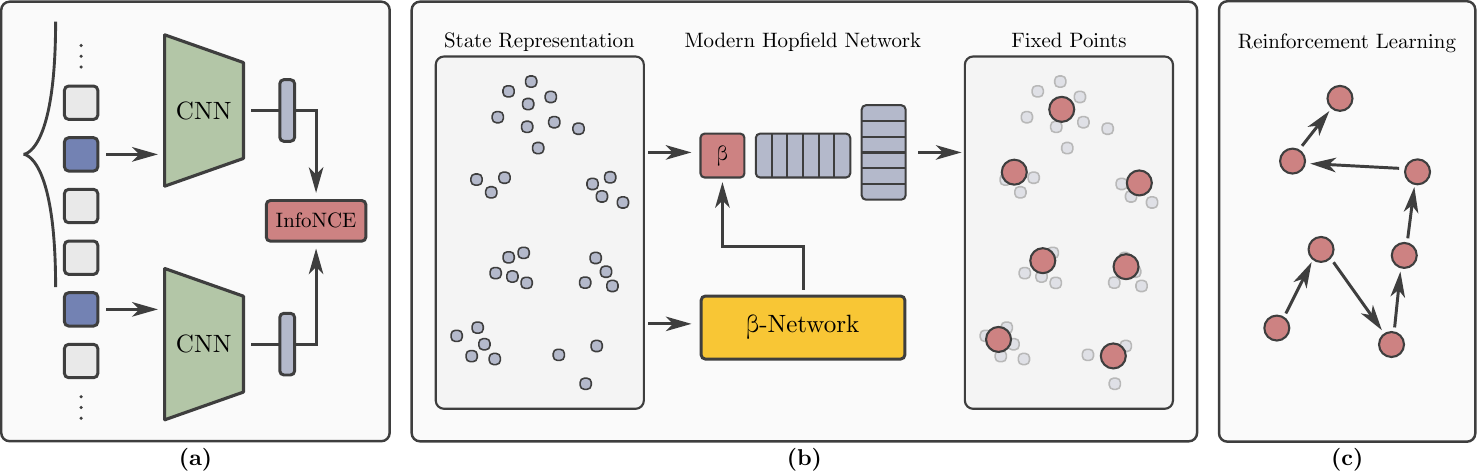}
    \caption{\textit{Contrastive abstraction learning.} \textbf{(a)} 
    The first phase applies self-supervised learning via contrastive learning to represent
    states with sequential proximity in a similar way. 
    Using the InfoNCE objective, two sequentially close states are forced to have a similar representation and 
    non-close states dis-similar representation.
    \textbf{(b)} 
    In the second phase, a modern Hopfield network maps similar
    representations to the same fixed point which constitutes an abstract state. 
    \textbf{(c)} In the last phase, downstream tasks are solved in the reduced state space.}
    \label{fig:overview}
\end{figure*}

\paragraph{Intuition for our approach.}
Typically, an agent remains in the same abstract state  
after a state transition, while leaving the abstract state is a rare event.
Such abstract states may be basic locations, achieved subgoals, current demands, 
inventory, or health conditions. 
If the agent is in the office and wants to buy from the
supermarket, then abstract states of the location might be 
office building, street, and supermarket. 
The first subgoal would be to leave the office building to the street,
the second subgoal to walk on the street to the supermarket, 
and the final subgoal to buy from the supermarket.
These abstract states may be identified by both sequential proximity 
of original states in the trajectories
and similarities of state representations.
Therefore, original states should be clustered to create abstract states.
In the next step, abstract actions have to be constructed that
allow to transit from one abstract state to another. 
These abstract actions can be subagents, like a subagent for 
leaving the building to the street.
Given abstract states, subagents can be readily trained via reinforcement learning.
A start-goal-subagent is placed in a state that belongs to the start abstract state and
receives a reward if it reaches a state that belongs to the goal abstract state.

We propose to learn abstractions, i.e.\ proper state representations,
through self-supervised learning.
States with sequential proximity in 
a trajectory are assumed to belong to the same abstract state.
Therefore, in the first phase we use contrastive learning to 
map states with sequential proximity to a similar representation.
In the second phase, similar representations are mapped by a modern Hopfield network (MHN)~\cite{Ramsauer:21}
to the same fixed point.
Such a fixed point constitutes an abstract state.
MHNs allow  
abstraction at different levels 
since a temperature parameter determines the number of fixed points.
Therefore, the same MHN with the same stored representations
enables finer or coarser abstractions. 
Further, we learn
the temperature parameter of the MHN.
This is useful as some tasks need to distinguish between states while in some cases we can unify them.
Our main contributions are:

\begin{itemize}
    \item We propose a novel method called "\textit{contrastive abstraction learning}" 
     to learn abstractions in state space without rewards using contrastive learning in the first phase
     and MHNs in the second phase. 
     Contrastive learning captures the structure of the environment by constructing state clusters, while
     MHNs supply the abstract states given state clusters. 
    \item \textit{Contrastive abstraction learning} controls the level of abstraction 
     by a temperature parameter of the MHN. 
     The control function adjusts the temperature based on the current state and is 
     trained by contrastive learning. 
    \item We study the resulting abstractions in different environments of varying complexity and 
    evaluate the performance of our \textit{contrastive abstraction learning} in downstream tasks. 
\end{itemize}

\section{Method}
We propose the novel \textit{contrastive abstraction learning}
to find abstract states for RL.
After learning, original states are mapped to the same abstract state 
if they are in sequential proximity in the training trajectories.
For example, in a grid world,
all states within a room
would be mapped to the same abstract state.
Therefore the number of states is reduced to the number of rooms.
Next, we have to construct a small number of sub-policies to move from one
room to another room.
Importantly, finding such an abstract state does not depend on a goal or reward 
and, therefore, can be used for various downstream RL tasks.
In the grid world example, representing the environment by the rooms helps
to solve any task where the agent has to visit specific rooms.

Figure~\ref{fig:overview} shows an overview of 
our \textit{contrastive abstraction learning}, which consists of three phases.
In the first phase, we sample two states from a sequence of states based 
on their sequential proximity and use contrastive learning to learn a representation that maps sequentially proximal states to a similar representation.
% The training sequences are trajectories, which may stem from experts, are prerecorded, 
% or are generated by a random policy.
The training trajectories might be expert examples, recorded examples,
or examples sampled from a random policy.
In the second phase, MHNs reduce 
the number of states to a small number of abstract states by mapping states to fixed points.
In order to control the level of abstraction, the temperature parameters of 
MHNs can be learned to be controlled.
In the final phase, a policy on this small set of abstract states is learned very 
efficiently since there are only few abstract states.
To learn the policy, we have to learn sub-policies that move the agent from one
abstract state to another. These sub-policies serve as actions for an MDP
built with the abstract states.

\subsection{Contrastive Learning of Sequential Proximal States}
\label{sec:method_contrastive}

With the advent of large corpora of unlabeled data
in vision and language, %, self-supervised learning via 
contrastive learning methods have become highly successful to learn expressive representations that can be adapted to various downstream tasks \cite{He:20moco,Chen:20,Radford:21,Fuerst:21, Siripurapu:22}.
Similarly, contrastive learning can be applied to data of trajectories to construct rich state representations.% useful for various downstream tasks.
We utilize contrastive learning to map states with sequential proximity to a similar representation.
Given an MDP, the resulting representation encodes the transition structure of the MDP.
For example, if an agent transits from state $s_1$ to $s_2$ after selecting an action $a$, 
then the representation of $s_1$ should be closer to $s_2$ than other states $s_t$.

We define our problem setting as a finite MDP without reward
to be a $3$-tuple of $(\sS, \sA, p)$ of finite set $\sS$ with states $s$, 
$\sA$ with actions $a$, 
transitions dynamics $p(s_{t+1} = s' \mid s_t = s, a_t = a)$.
Here, $t$ is
the position in a state-action trajectory $(s_1,a_1,\ldots,s_T,a_T)$ of length $T$.
The agent selects actions $a \sim \pi(s_t = s)$ based on the policy $\pi$, 
which depends on the current state $s$.
For contrastive learning,
pairs of states $(s_t,s_{t+k})$ from the same trajectory 
with small absolute value of $k$ are selected as positive pairs with high probability, while
other pairs are negative pairs, which also include states from different trajectories.
We sample a set of positive pairs for each mini-batch of trajectories in the following way. 
First, we uniformly sample a time index $t_i$ from a trajectory $\tau_{0:T}$.
Then, we sample the time index $t_j$ from the same trajectory
e.g.\ by centering a Gauss or Laplace distribution on $t_i$.
The Laplace distribution can be approximated by using a discount factor and normalizing the discounts, 
where states farther away from $i$ are more discounted and, therefore, have lower probability.
Hence, the samples of a positive pair are temporally related.
%Constrastive learning represents states similarly, when they often occur together. 
Negative pairs are samples from different trajectories or those with a large sequential distance.
With $N$ as the number of samples in the mini-batch, our contrastive objective \cite{vanDenOord:18} is
\begin{align}
  \label{eq:lossNCE}
  \rL_{\mathrm{InfoNCE}} \ &= \ - \ \frac{1}{N}  \sum_{i=1}^N \ \ln \frac{\exp (\tau^{-1} \ s_{t_i}^T s_{t_j})} {\sum_{k=1}^N \exp (\tau^{-1} \ s_{t_i}^T s_{t_k})} \ .
\end{align}

\subsection{Sampling a Positive Pair}
\label{subsec:sampling_stratgies}
The sampling strategy to construct a positive pair for the contrastive objective 
has a profound impact on the learned representation.
Given a trajectory, the timestep $t_i$ of the first state of a positive pair is sampled uniformly. 
Then, we center a distribution at the timestep $t_i$ and then sample the timestep $t_j$ of the second state to complete the positive pair. 
We explore the following four distributions located at $t_i$ for sampling $t_j$: 
(I) uniform, (II) Gaussian, (III) Laplace, and (IV) exponential.
With sampling from a uniform distribution, 
the second timestep $t_j$ is sampled uniformly from the same trajectory, 
thus not using sequential proximity.
For sampling from a Gaussian distribution, 
we center a Gaussian with standard deviation $\sigma$ at $t_i$ 
and select $t_j$ that is closest to a sample that can be from both the future and the past.
Analog for sampling from a Laplace distribution with scale parameter $b$.
Also analog for sampling from an exponential distribution inverse scale $\gamma$.
Since only positive values are sampled, only future states are sampled.
For additional details on sampling strategies see Appendix Sec.~(\ref{app:method_contrastive}).

\subsection{Abstractions via Modern Hopfield Networks}
\label{sec:method_abstraction_mhn}

After learning state representations based on sequential proximity, we reduce their number by mapping them to abstract states.
With a small state space RL can efficiently solve an MDP for a given reward function. 
MHNs \citep{Ramsauer:21} map states to fixed points that 
correspond to abstract states.
Hopfield networks, introduced in the 1980s, are binary associative memories 
that played a significant role in popularizing artificial neural networks \citep{Amari:72,Hopfield:82,Hopfield:84}. 
These networks were designed to store and retrieve samples~\citep{Chen:86,Psaltis:86,Baldi:87,Gardner:87,Abbott:87,Horn:88,Caputo:02,Krotov:16}.
In contrast to traditional binary memory networks, 
we employ continuous associative memory networks with exceptionally high storage capacity. 
These MHNs, utilized in deep learning architectures, 
have an energy function with continuous states and 
possess the ability to retrieve samples with just a single update \citep{Ramsauer:21}. 
The update rule of MHNs is guaranteed to converge to a fixed point. 
These fixed points correspond to attractors in the network's energy landscape, and the network dynamics converge towards these attractors when starting at an input pattern.
MHNs have already been demonstrated to be successful 
in diverse fields~\cite{Widrich:20, Seidl:22, Widrich:21, Paischer:22, Fuerst:21}.
In summary, MHNs store and retrieve patterns similar to the query (the input).

We leverage the mechanism of MHNs 
to map every input pattern to a fixed point.
The MHNs have $N$ state representations $\Bu_i$
as stored patterns $\BU = \left( \Bu_1, \ldots, \Bu_N \right)$. 
The update rule of MHNs with inverse temperature $\beta$ is
\begin{align}
\label{eq:hopfield_update}
\Bxi^{t+1} \ &= \ \BU \ \soft ( \beta \BU^T \Bxi^t) \ ,
\end{align}
which is applied until convergence.
The query $\Bx$ of the MHN is a state representation and
serves as initial state $\Bxi^0 = \Bx$.
To determine all fixed points of the MHN,
we query with the $i$th stored representations $\Bxi^0 = \Bu_i$.
Therefore, we obtain $N$ fixed points of the MHNs.
After removing duplicates, we get all 
$M$ different fixed points $\left( \By_1, \ldots, \By_M \right)$
of the MHN that stores $\BU$.

By adjusting the inverse temperature $\beta$, 
we can control the number of fixed points $\By$,
and thus the level of abstraction.
For $\beta=0$ there is only one fixed point $\By$,
which is the average of all stored patterns. 
Conversely, for large $\beta$ the stored pattern is
retrieved that is closest to the query.
For intermediate $\beta$ values, we obtain meta-stable states,
that means several stored patterns converge to a fixed point they share.
The larger $\beta$ the smaller the meta-stable states, i.e.\ less stored 
patterns are mapped to the same fixed point.
Thus, small values for $\beta$ result in a small number 
of fixed points while high values of $\beta$ in a larger number of fixed points.
Therefore, adjusting $\beta$ to control the number of fixed points, 
and thus the granularity of our abstract representation, is crucial.

\subsection{Controlling the Level of Abstraction}
\label{sec:method_learning_beta}
For a given state $s$, we first obtain its state representation $\Bx$,
which serves as input to the MHN. 
Then we obtain the fixed point $\By=\By(\Bx)$, which is the abstract state.
Controlling the level of abstraction is essential to generate representations that are 
useful for a large number of tasks.
MHNs are well suited to achieve this control via the
inverse temperature $\beta$.

\paragraph{Training a $\beta$-network.} 
To determine a suitable $\beta$ for a given state representation $\Bx$ we train a separate $\beta$-network (see overview Figure~(\ref{fig:overview})).
%This output is the inverse temperature of the modern Hopfield network. 
This $\beta$-network is trained using a contrastive setup.
%in an end-to-end fashion.
%
One branch of the contrastive model applies random dropout to a given state representation $\Bx$ to obtain a masked representation $\Bx'$.
By masking features in the input, the network is forced to adjust the level of abstraction in such a  way, 
that even masked representations are assigned to the correct fixed point. 
A second branch consists of the $\beta$-network followed by a MHN. 
The original $\Bx$ is input to the $\beta$-network, a small FNN, which regresses a $\beta$ value. 
This $\beta$ value is then used in the MHN to map the state representation $\Bx$ to a fixed point $\By$.
The output of the two branches, the masked representation $\Bx'$ and the fixed point $\By$, form a positive pair. 
Other masked representations from the mini-batch serve as negatives. 
The contrastive model is trained with the InfoNCE objective. See Appendix Sec.~(\ref{app:learned_beta}).

\begin{figure*}[t]
    \centering
    \includegraphics[width=\linewidth]{figures/sampling_analysis_6_v2.pdf}
    \caption{Visualization of learned representation of the states for 
    different sampling techniques for contrastive learning. 
    \textbf{Left:} Two trajectories ($\mathcolor{calRed}{\star}$ and 
    $\mathcolor{calGreen}{\blacktriangle}$) in the Maze2D environment. 
    \textbf{Right:}  Different sampling techniques of positive pairs for contrastive learning.
    Red points are states in the dataset, while blue and green are the trajectories from left.
    Sampling from the Laplace distribution leads to representation well suited for 
    abstraction, while Gaussian does not yield clear clusters.
      }
    \label{fig:sampling}
\end{figure*}

\subsection{Using Abstraction for Downstream Tasks}

Abstraction of states can be a powerful tool for improving the efficiency and effectiveness 
of planning and RL tasks. 
Abstract states can be used in several ways for downstream applications.
The following three approaches demonstrate ways in which abstraction can be applied to downstream tasks. 
Each offers unique advantages in terms of reduced state space and efficient learning. 
The specific approach can be chosen depending on the characteristics of the task and the available resources.

\paragraph{Learning a policy from abstract states (Abstract Policy).} 
By reducing the state space to a set of abstract states, we also reduce the complexity of the problem. 
Instead of operating in the original state space, the policy can now focus on making decisions based on abstract states. 
We refer this approach as \textit{Abstract Policy}.
For this approach, we assume that the optimal policy selects the same action for all original states mapping to one abstract state.
While this is a strong assumption we show experimentally, that this approach is viable.
% When this assumption is fulfilled learning a policy from abstract states is much more efficient than in the original state space.

\paragraph{Learning sub-policies as meta-actions (Meta Policy).}
Sub-policies allow the agent to transition from one abstract state to another, thereby acting as meta-actions. 
Such sub-policies can be learned from existing trajectories or newly generated trajectories using imitation learning or other RL algorithms.
For example, the agent can learn to imitate trajectories between abstract states via Behavior Cloning (BC). 
The sub-policy for the transition from abstract state $\By_1$ to $\By_2$ starts in a state mapped to $\By_1$ and 
receives a reward if it arrives at a state which maps to $\By_2$. 
Subsequently, a meta-policy is trained to select the appropriate sub-policy based on the current abstract state. 
The advantage of this approach is that it focuses on learning and selecting from the small abstract state space, leading to faster and more targeted learning. 
We refer to this approach as \textit{Meta Policy}. 

\paragraph{Goal-conditioned planning from a graph over abstract states (Planning).}
In a third approach, a graph is constructed over the abstract states, representing their relationships and connectivity. 
This graph can then be utilized for planning and decision-making. 
Given a goal state, the agent can plan a sequence of abstract states that lead to the desired outcome using graph search algorithms. 
Additionally, learned sub-policies can be employed to execute actions within the environment, facilitating the agent's progress towards the goal state. 
This approach combines the benefits of abstraction, graph-based planning, and learned sub-policies to enable effective goal-directed behavior.

% {color{blue} TODO:7: Write how these abstract fixed points can be used in Reinforcement Learning or sequential decision-making}

\section{Experiments}

In order to show the effectiveness of \textit{contrastive abstraction learning} we perform a series of experiments.
First, we verify that abstract representations encode the structure of the environment.
Using trajectories collected from a diverse set of environments we visualize the abstract representations via fixed points.
Next, we investigate the connection between $\beta$ and the level of abstraction.
Finally, we show experiments for each of the three approaches of using \textit{contrastive abstraction learning} for downstream tasks detailed in the previous section.

\begin{figure*}[t]
    \centering
    \includegraphics[width=0.9\linewidth]{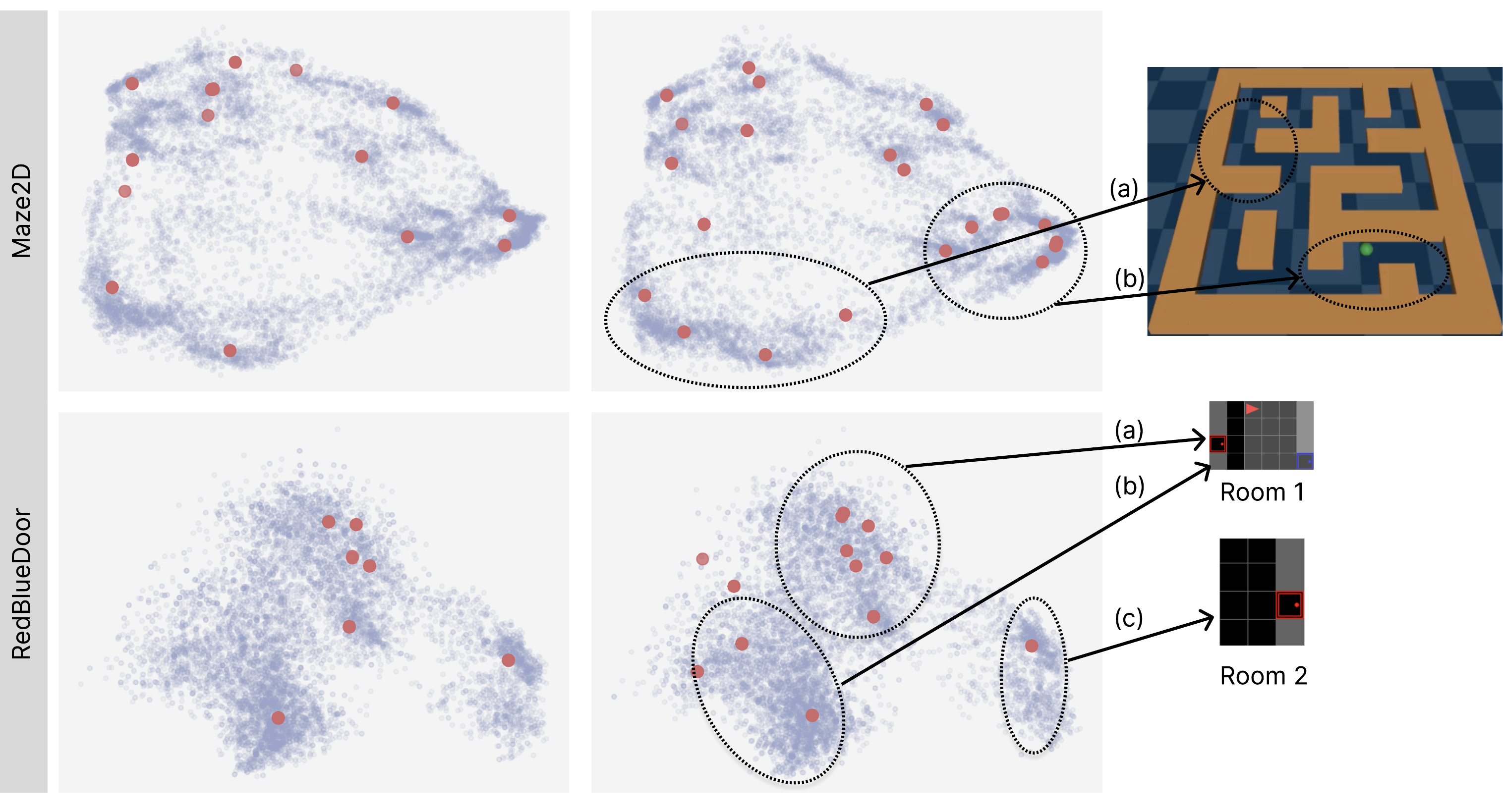}
    \caption{Learned representations (blue) and fixed points (red) 
    of a MHN for Maze2D and RedBlueDoor. 
    The parameter $\beta$ of the MHN increases from left to right, 
    thus increasing the number of fixed points.
    The learned representation forms clusters of states where corridors (top) or rooms (bottom) are close.
    These clusters are mapped to abstract states via an MHN. In the rooms example (bottom row), (a) corresponds to states where the agent is in Room 1 and the red door is closed, (b) corresponds to the red door being open, (c) corresponds to agent being in Room 2.} 
    \label{fig:increasing_beta}
\end{figure*}

\subsection{Environments}
We perform all experiments on a set of environments with increasing complexity selected or designed to showcase the effectiveness of a learned abstract representation.
Furthermore, for learning a contrastive representation as well as for learning the abstract representation we require a dataset of trajectories.
Therefore, we select a set of trajectories for each environment from existing datasets if available or by sampling from a random policy in case no dataset is available.
% Therefore, we select a set of trajectories for each environment with a random policy, unless stated otherwise.
Alternatively, a dataset can easily be acquired with an iterative approach by interacting with the environment but we found that the representation learned from randomly sampled trajectories is often sufficient.

\textbf{(I)} \textit{CifarEnv} is an environment with images and classes from the CIFAR-100 dataset. It is used to verify if the method obtains abstract representations. We select 10 classes with defined transition structures for various goals and tasks. The agent remains in the same state for a specified number of timesteps (e.g., 8), receiving images from the corresponding class. After the timesteps, the agent selects an action to transition to the next state. We sample 100,000 timesteps from this environment to create a dataset for contrastive abstraction learning.
\textbf{(II)} \textit{Maze2D-large} is an environment from Mujoco Suite \cite{Todorov:2012:mujoco}. 
A robot must navigate to a given goal state from a random starting position.
We use trajectories from the D4RL dataset \cite{Fu:2020:d4rl} for our experiments.
\textbf{(III)} \textit{RedBlueDoor} is part of the Minigrid library \cite{Chevalier:2018:minigrid} and is a partially observable environment.
We generate a dataset by storing all interactions during training of a policy with Proximal Policy Optimization (PPO) \cite{Schulman:17}.
We use the environment and the dataset of human demonstrations released for the MineRL competition \cite{Guss:19comp}.

\begin{figure*}[t]
    \centering
    \subcaptionbox{\label{sfig:a}Fixed beta}{\includegraphics[width=0.4\linewidth]{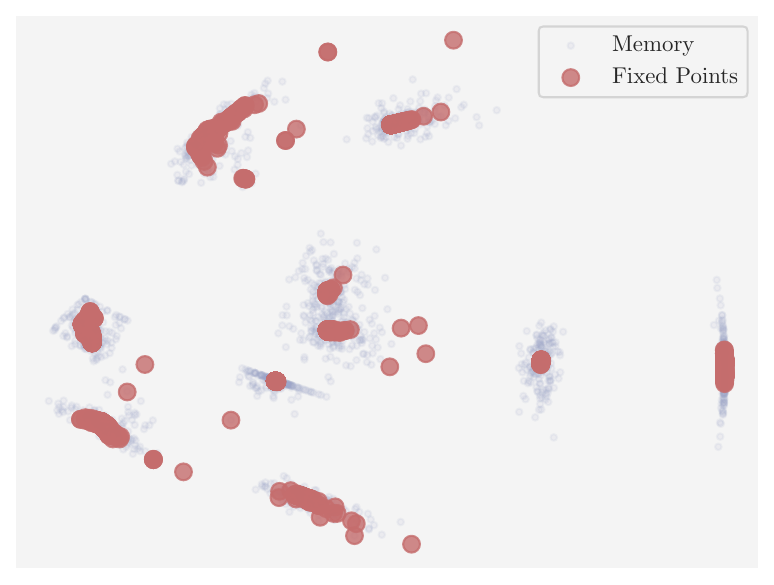}}
    \hspace{0.65cm}
    \subcaptionbox{\label{sfig:b}Learned beta}{\includegraphics[width=0.4\linewidth]{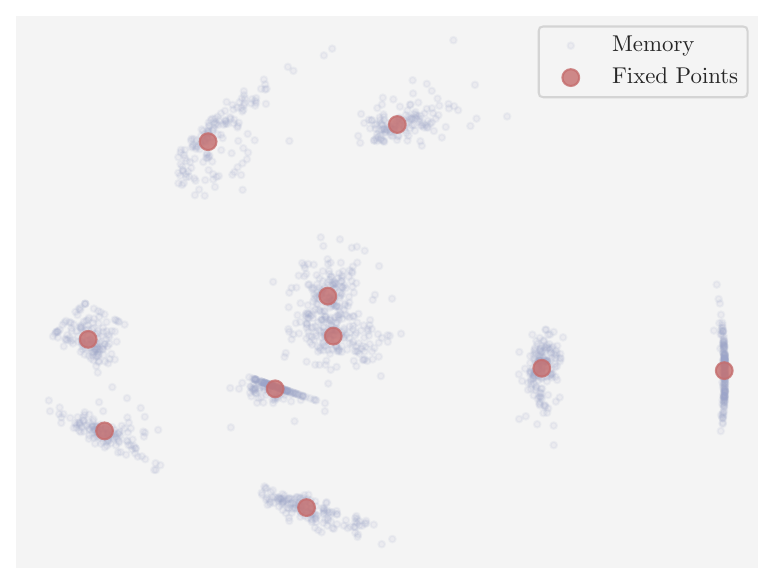}}    
    \caption{Visualization of all states (blue) in the memory, with the resulting fixed points (red) in \textit{CifarEnv}. 
    We compare the number of fixed points when the temperature parameter is fixed (a) and when the temperature parameter is learned (b). The learned temperature parameter reduces the state space to a number of fixed points equal to the number of Cifar classes in the environment. } 
    \label{fig:learned_beta}
\end{figure*}

\subsection{Contrastive and Abstract Representation}
In order to show that \textit{contrastive abstraction learning} results in a representation that encodes the structure of the environment we present 
qualitative analysis for the various environments.
We visualize the learned representation after contrastive learning and the abstract representation obtained via fixed points from MHNs.
We use PCA to downproject both the representation after contrastive pre-training as well as the abstract representation.

\paragraph{Sampling strategies for contrastive pre-training.} We visualize the learned representation after contrastive learning of 
sequentially proximal states for different sampling strategies (\ref{subsec:sampling_stratgies}) in Figure \ref{fig:sampling}. 
We visualize all samples after contrastive pre-training and highlight two trajectories.
While on first look the representation obtained when sampling positive pairs using the Gaussian strategy seems good, for subsequent abstraction this strategy is not well suited.
The representation almost perfectly mirrors the original state space, therefore similar states are not closer to each other than without contrastive pre-training.
As expected, the Uniform strategy results in unclear clusters and thus a sub-par representation.
From the remaining two we prefer sampling with the Laplace strategy as clusters of states emerge and states where both trajectories overlap are closer in the representation.

% Note Markus: we should show "increasing beta" and "learned beta" on the same or at least an overlapping set of environments
\paragraph{Controlling the level of abstraction through $\beta$.} Next, we visualize abstract states obtained via fixed points of the MHN.
First, we show the effect of $\beta$ on the number of abstract states.
In Figure \ref{fig:increasing_beta} we show three different settings of beta for three environments, namely \textit{Maze2D-large} and \textit{RedBlueDoor}.
From left to right we increase $\beta$ and therefore the number of abstract states that emerge.
This shows that finding a suitable value for $\beta$ is crucial for solving downstream tasks.

\paragraph{Learning $\beta$.} Next, we show on \textit{CifarEnv} that learning $\beta$ works well and results in a set of abstract states that is both small and covers all modes of the state space.
Figure \ref{fig:learned_beta} (a) shows a manually selected value for $\beta$ that at first glance results in a good abstract representation.
However, the set of abstract states is still much larger than necessary.
In Figure \ref{fig:learned_beta} (b) the abstract states, represented by fixed points of the MHN, are much sparser while still covering the state space.
Manually tuning $\beta$ for each environment would be time-consuming, therefore learning an good value for $\beta$ is crucial.

\begin{figure*}[t]
    \centering
    \includegraphics[width=1.0\linewidth]{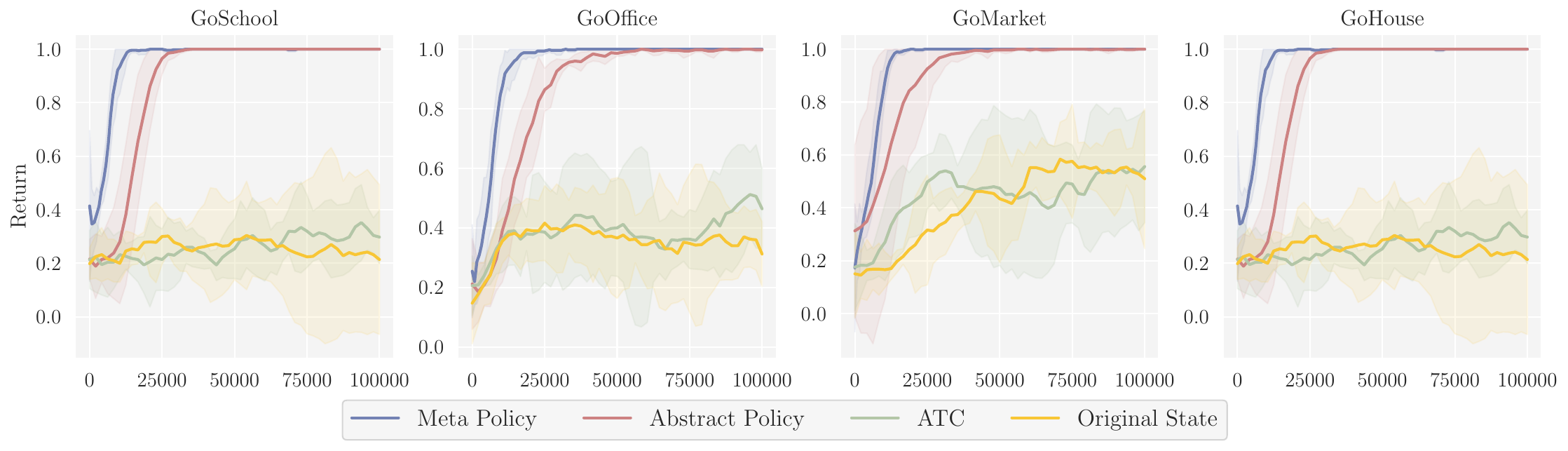}
    \caption{We show training of policies from abstract states to solve four tasks in \textit{CifarEnv}. 
    \textit{Abstract Policy} uses abstract states but the original action space. 
    \textit{Meta Policy} selects actions from an abstract action space and executes sub-policies.
    \textit{Contrastive Representation} uses state representations from contrastive pre-training.
    For \textit{Original State} both action and state space are not changed.
    Both methods using abstract states can solve all tasks.}
    \label{fig:cifar_ppo}
\end{figure*}

\subsection{Abstract Policy}

The first approach for using the abstract representation for downstream tasks we investigate is training an \textit{Abstract Policy}.
We assume that the optimal policy selects the same action for every state belonging to an abstract state. %original state mapping to an abstract state when transitioning from one abstract state to another.
As this assumption holds in the \textit{CifarEnv} we train a policy using PPO. 
At every timestep $t$ the original state $s_t$ is first mapped to it's corresponding abstract state $\By$.
This abstract state is then used by the policy network.
The original action space is not changed.
Figure \ref{fig:cifar_ppo} shows that \textit{Abstract Policy} can solve all tasks in this environment, only outperformed by the next \textit{Meta Policy}.
For comparison we include two additional baselines.
\textit{Contrastive Representation} is trained with PPO using the representation after contrastive pre-training, thus without abstract states.
\textit{Original State} is a PPO baseline on the original state space without constrastive abstraction learning.

\subsection{Meta Policy}

Next, we train the \textit{Meta Policy}.
In contrast to the \textit{Abstract Policy}, the original action space is replaced by an abstract action space where each action represents a transition from one abstract state to another.
These abstract actions are learned, either by interacting with the environment or, as in our case, via BC.
Thus, each abstract action is a sub-policy that imitates behavior from the dataset.
When an abstract action is selected, the corresponding sub-policy is executed in the environment.
The \textit{Meta Policy} is again trained with PPO.
In Figure \ref{fig:cifar_ppo} we show results using this approach in the \textit{CifarEnv}.
\textit{Meta Policy} outperforms all other approaches and baselines.

\subsection{Planning}

\begin{figure*}[t]
    \centering    
    \subcaptionbox{\label{sfig:testa}Original graph (\textit{CifarEnv})}{\includegraphics[width=0.4\linewidth]{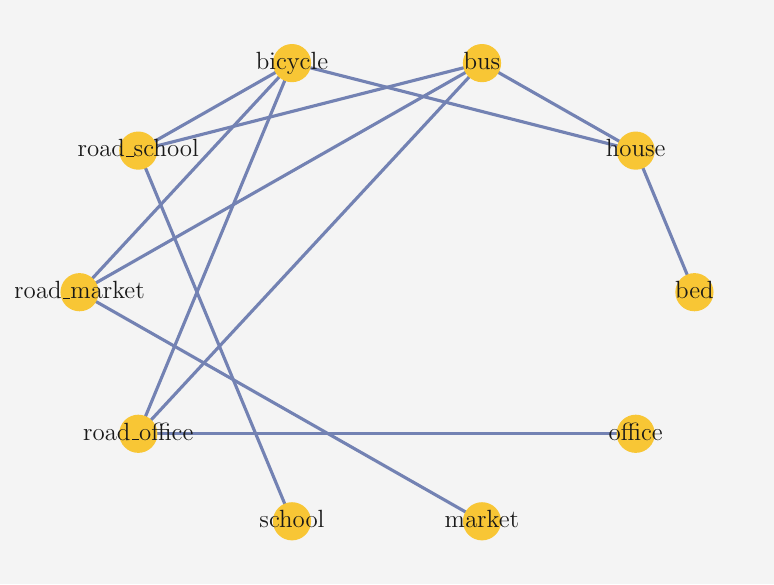}}
    \hspace{0.65cm}
    \subcaptionbox{\label{sfig:testb}Graph with fixed points (\textit{CifarEnv})}{\includegraphics[width=0.4\linewidth]{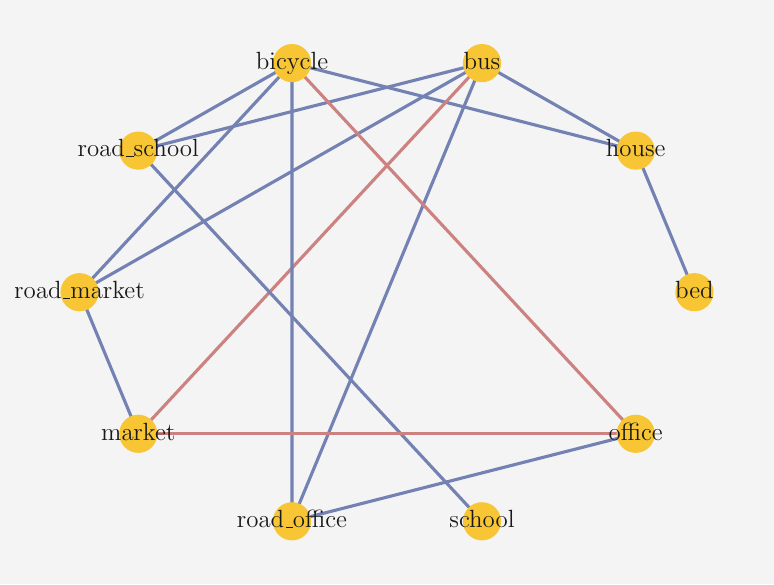}}
    \caption{We compare the actual underlying transition graph (a) between states and the graph we construct using fixed points (b). Such a graph can be used for high-level planning in the environment.} 
    \label{fig:cifar_planning}
\end{figure*}

The third approach for obtaining a policy from abstract states is \textit{Planning}.
Here, we extract a graph from the abstract states describing an environment and then use it for planning. 
The graph is constructed by calculating the cosine similarity of all abstract states with each other.
Then, a threshold is applied to remove connections between dissimilar abstract states.
In Figure \ref{fig:cifar_planning} we compare the true graph and the graph constructed from the abstract states.
% (a) shows the true graph we used to construct the environment.
% In Figure \ref{fig:cifar_planning} (b) we show a graph constructed from abstract states after learning.
While there is a some additional edges, the constructed graph is sufficient to solve almost all tasks.
In this experiment, we randomly sample goal states and plan a path towards it using the graph. We find the shortest path from the current abstract state to the abstract state of the goal on the graph. 
% In this experiment, we randomly sample goals from abstract states in \textit{CifarEnv}.
% We then use the graph we constructed from the abstract states and use XXX to obtain a policy to reach this goal.
Similar to \textit{Meta Policy}, we use sub-policies to transition from one abstract state to another.
These are again learned with BC from the dataset.

\begin{table}[h]

\caption{Planning over graphs. For each environment 1000 abstract start and goal states are sampled.
A graph is constructed and used for planning a policy to transition from start to goal state. 
We show the rate of success of reaching the goal state when the plan is executed in the environment.}
\vspace{0.7em}
\centering
\begin{tabular}{lccc}
    \toprule
    Environment  & CifarEnv &  Maze2D-Large & Maze2D-Medium \\
    \midrule
    Success (\%) & 86       &   88          &  92     \\
    \bottomrule
\end{tabular}
\label{tab:cifar_planning}
\end{table}

In addition to \textit{CifarEnv} we evaluate the robustness of constructing a valid graph from abstract states. 
For each environment in Table \ref{tab:cifar_planning} we sample 1000 start and goal states. Using a graph, we find a plan to transition from the start to goal state and execute this plan in the environment. 
% Then, we construct the graph and a plan to transition from start to goal state and execute this plan in the environment.
We then record if the goal state is reached within a certain number of timesteps. For more details see Appendix Sec.~(\ref{app:experiments}).

\section{Related Work}
%\textbf{Related Work.}
Recently, contrastive learning was very successful at
constructing powerful representations with clusters.
In the embedding space, contrastive learning brings 
representations of paired inputs closer to one another while moving unpaired 
inputs farther away from each other \cite{Chen:20, Radford:21, Rombach:22}. Following this success, Contrastive Learning has been used in RL to learn representations \cite{Srinivas:20, Eysenbach:22}. 
CURL~\cite{Srinivas:20} learns representations on images by pairing a state with its augmented version. 
CoBerl~\cite{Banino:22} masks elements of trajectories in the input and then tries to predict them, analogously to BERT in the language domain.
%SEPP is this contrastive? It is self-supervised. Maybe move to self-supervised.
\cite{Eysenbach:22} pairs state-actions with future state-actions in a trajectory.
Contrastive learning has been used as unsupervised pretraining for RL. 
APS~\cite{Hao1:21} and ATS~\cite{Hao2:21} learn representations by maximizing entropy during an 
unsupervised phase and fine-tuning for a subsequent task. 
%SEPP is this contrastive?
ATC \cite{Stooke:21} pairs state with a state that is $k$ steps ahead in the trajectory. 
SPR~\cite{Schwarzer:20, Schwarzer:21}, KSL~\cite{Mcinroe:21} and \cite{Anand:21} predict the representation of $k$ future steps and bring it closer to the actual representation.
Proto-RL~\cite{Yarats:21} tries similarly to use ideas from Swav~\cite{Caron:20} and offline RL \cite{Schweighofer:21, Schweighofer:22} to first explore the environment in 
pre-training to learn representation from diverse samples and then fine-tunes on downstream task.

Abstraction in RL is a well-studied area \cite{Bertsekas:89, Givan:03, Ravindran:03a, Ravindran:03b, Sutton:99, Li:06, Daniel:16, Bacon:17, Vezhnevets:17, Kulkarni:16, patil:22, Patil:22b, Caggiano:22}. 
Bisimulation \citep{Givan:03, Ferns:12, Castro:19, Zhang:20} tries to simplify MDPs by identifying states which lead to similar behaviors (bisimilar states). 
Methods based on bisimulation learn abstractions that are dependent on a particular reward function.%, while %contrastive abstraction tries to learn abstraction independent of a reward function.  
\ \textit{Contrastive abstraction learning} enables the identification of sub-goals and sub-tasks, thus making it relevant to hierarchical reinforcement learning (HRL) approaches such as the option framework \cite{Sutton:99}, the MAXQ framework \cite{Dietterich:00}, and the recursive composition of option models \cite{Silver:12}. 
Several approaches have been proposed to tackle the problem of learning good options. 
One approach is selecting frequently observed solution states as targets \cite{Stolle:02}. Another strategy involves employing gradient-based techniques to enhance the termination function for options \cite{Comanici:10, Mankowitz:16, Levy:12}. 
\cite{Levy:12} employed policy gradient optimization to learn a unified policy comprising intra-option policies, option termination conditions, and an option selection policy. 

Parametrized options can be learned by treating the termination functions as hidden variables and applying expectation maximization \cite{Daniel:16}. Intrinsic rewards have also been utilized to learn policies within options, while extrinsic rewards are used to learn the policy over options \cite{Kulkarni:16}. \cite{Bacon:17} proposed a method that jointly learns options and their associated policies using the policy gradient theorem. Additionally, a manager module operating on a slow time scale has been introduced to learn sub-goals, which are subsequently achieved by a worker module operating on a fast time scale \cite{Vezhnevets:17}.

\section{Conclusion}
We propose \textit{contrastive abstraction learning}, a novel approach for state abstraction in RL.
Abstract states drastically reduce the number of states for planning in RL.
We use contrastive learning to learn a representation where states in sequential proximity are mapped to similar representations.
Then we use MHNs to determine abstract states via fixed points. 
Furthermore, we can adjust the level of abstraction by controlling the number of fixed points. 
We show how to learn the level of abstraction via contrastive learning.
We propose three approaches for using such an abstract representation in RL for efficiently learning policies.
Our experiments show, that our approach is able to find a small set of abstract states in several diverse environments and that 
learning policies based on this representation can be efficient.

\begin{ack}
The ELLIS Unit Linz, the LIT AI Lab, the Institute for Machine Learning, are supported by the Federal State Upper Austria. We thank the projects AI-MOTION (LIT-2018-6-YOU-212), DeepFlood (LIT-2019-8-YOU-213), Medical Cognitive Computing Center (MC3), INCONTROL-RL (FFG-881064), PRIMAL (FFG-873979), S3AI (FFG-872172), DL for GranularFlow (FFG-871302), EPILEPSIA (FFG-892171), AIRI FG 9-N (FWF-36284, FWF-36235), AI4GreenHeatingGrids(FFG- 899943), INTEGRATE (FFG-892418), ELISE (H2020-ICT-2019-3 ID: 951847), Stars4Waters (HORIZON-CL6-2021-CLIMATE-01-01). We thank Audi.JKU Deep Learning Center, TGW LOGISTICS GROUP GMBH, Silicon Austria Labs (SAL), FILL Gesellschaft mbH, Anyline GmbH, Google, ZF Friedrichshafen AG, Robert Bosch GmbH, UCB Biopharma SRL, Merck Healthcare KGaA, Verbund AG, GLS (Univ. Waterloo) Software Competence Center Hagenberg GmbH, T\"{U}V Austria, Frauscher Sensonic, TRUMPF and the NVIDIA Corporation.    
\end{ack}

\bibliography{minimo}
\bibliographystyle{plainnat}

%%%%%%%%%%%%%%%%%%%%%%%%%%%%%%%%%%%%%%%%%%%%%%%%%%%%%%%%%%%%
% \newpage
% \section{Supplementary Material}

\onecolumn
\appendix

\setcounter{theorem}{0}
\setcounter{definition}{0}
\setcounter{table}{0}
\setcounter{figure}{0}
\setcounter{equation}{0}

\renewcommand{\thefigure}{A\arabic{figure}}
\renewcommand{\thetable}{A\arabic{table}}
\renewcommand{\theequation}{A\arabic{equation}}

\resumetocwriting
\setcounter{tocdepth}{3}

\renewcommand{\contentsname}{\large Contents of the Appendix}
\tableofcontents
\newpage

\section{Extended Related Work}

\subsection{Contrastive Learning}

Non-Contrastive learning objectives for self-supervised learning, such as 
those of BYOL \citep{Grill:20} and SimSiam  \citep{Chen:21simsiam}
do not require negative samples.
However, most self-supervised learning methods use contrastive learning.
The most popular constrative objective is InfoNCE \citep{vanDenOord:18}. 
InfoNCE pairs an anchor sample with a positive sample (positive pair) and then contrasts this pair to 
pairs of the anchor sample with negative samples (negative pairs). 
InfoNCE has been utilized in transfer learning \citep{Henaff:19}, 
aiding natural language response suggestions \citep{Henderson:17}, 
acquiring sentence representations from unlabelled data \citep{Logeswaran:18}, 
and facilitating unsupervised feature learning by maximizing distinctions between instances \citep{Wu:18}. 
Furthermore, InfoNCE has been effectively employed for learning 
visual representations in Pretext-Invariant Representation Learning (PIRL) \citep{Misra:20}, 
Momentum Contrast (MoCo) \citep{He:20moco}, and SimCLR \citep{Chen:20}.
Given the success of the InfoNCE \cite{vanDenOord:18} with sequential data \cite{Henderson:17, Logeswaran:18},
we use it as an objective for contrastive learning. 
Specifically, we use contrastive learning using the InfoNCE \citep{vanDenOord:18} objective.

Using contrastive learning for constructing powerful representations
is well established.
Contrastive Predictive Coding (CPC)  \citep{vanDenOord:18}
learns abstract representations in an unsupervised way.
The representation build by CPC was general enough that it allowed for 
transfer learning \citep{Henaff:19}.
Contrastive learning has been applied to natural language tasks
to learn sentence representations from unlabelled data \citep{Henderson:17,Logeswaran:18} 
and to unsupervised feature learning \citep{Wu:18}.
Also, SimCSE learns sentence embeddings \citep{Gao:21}
In vision, 
Pretext-Invariant Representation Learning (PIRL)
contrasts representations of transformed versions of the same image with 
representations of other images \citep{Misra:20}.
Momentum Contrast (MoCo) was very successful for unsupervised
visual representation learning \citep{He:20moco}.
SimCLR is one of the best-known 
contrastive learning methods for visual representations,
which was highly effective for transfer learning \citep{Chen:20}. 
Vision applications have been extended to video representation learning \citep{Han:20}.
A milestone in contrastive learning was 
Contrastive Language-Image Pre-training (CLIP), which
yielded very impressive results at
zero-shot transfer learning \citep{Radford:21,Radford:21arxiv}.
CLIP learns expressive image embeddings directly from raw text, thereby
leverages a much richer source of supervision than just labels.
The representation learned by CLIP are so powerful that 
a plethora of follow-up publications used CLIP's representations.
The CLIP model is used in Vision-and-Language tasks \citep{Shen:21}. 
The CLIP model guided generative models via an additional training objective \citep{Bau:21,Galatolo:21,Frans:21} and improved clustering of latent representations \citep{Pakhomov:21}.
It is used in studies of out of distribution performance \citep{Devillers:21,Milbich:21,Miller:21}, 
of fine-tuning robustness \citep{Wortsman:21},  of zero-shot prompts \citep{Zhou:21} 
and of adversarial attacks to uncurated datasets \citep{Carlini:21}. 
It stirred discussions about more holistic evaluation schemes in computer vision \citep{Agarwal:21}. 
Multiple methods utilize the CLIP model in a straightforward way 
to perform text-to-video retrieval \citep{Fang:21,Luo:21,Narasimhan:21}.
Latent diffusion models \cite{Rombach:22} lead to 
stable diffusion models version 1 \cite{StableDiffusion1:22} and version 2 \cite{StableDiffusion1:22}, 
which are a recent open-source image generation models 
comparable to proprietary models such as DALL-E 2 \cite{Ramesh:22}.
Stable diffusion model version 1 uses CLIP for textual guiding, while 
version 2 uses OpenCLIP. The representation learned by CLIP is key to gernerate
impressive images by stable diffusion models.
Stable diffusion models hat high impact on the scientific community as well as in the public.

\subsection{Contrastive Learning in Reinforcement Learning}
CURL\cite{Srinivas:20} learns representation on raw pixels by bringing closer the representation of a state and its augmented version. 
CoBerl\cite{Banino:22} uses contrastive learning to learn representations by using the masked prediction objective of Bert in the time domain for reinforcement learning.
\cite{Eysenbach:22} attempt to move the representation of a state-action closer to another state if it is present in the future and away from another state if it is not present in the future.
Some methods use contrastive learning during an unsupervised phase before online learning. 
APS \cite{Hao1:21} and ATS \cite{Hao2:21} learn representations by maximizing entropy during an unsupervised phase and fine-tuning for a subsequent task. 
ATC \cite{Stooke:21} brings closer the representation of a state to another state which is $k$ steps in the future. 
SPR \cite{Schwarzer:20, Schwarzer:21}, KSL \cite{Mcinroe:21} and \cite{Anand:21} predict the representation of $k$ future steps and bring it closer to the actual representation.
DreamerPro \cite{Deng:21} combines the ideas from Dreamer \cite{Hafner:19} and 
Swav \cite{Caron:20} to do Model-Based Reinforcement learning without reconstruction. 
Proto-RL \cite{Yarats:21} tries similarly to use ideas from Swav to first explore the environment in 
pre-training to lean representation from diverse samples and then fine-tunes on downstream task. 
Palm Up \cite{Liu:22} connect large-scale vision models to reinforcement learning by using them as an environment to learn representations. 

\subsection{Abstraction in Reinforcement Learning}

Abstraction in RL is a well-studied area \cite{Bertsekas:89, Givan:03, Ravindran:03a, Ravindran:03b, Sutton:99, Li:06, Daniel:16, Bacon:17, Vezhnevets:17, Kulkarni:16}. 
Bisimulation \citep{Givan:03, Ferns:12, Castro:19, Zhang:20} tries to simplify MDPs by identifying states which lead to similar behaviors (bisimilar states). 
Methods based on bisimulation learn abstractions that are dependent on a particular reward function.%, while %contrastive abstraction tries to learn abstraction independent of a reward function.  
\ \textit{Contrastive abstraction learning} enables the identification of sub-goals and sub-tasks, thus making it relevant to hierarchical reinforcement learning (HRL) approaches such as the option framework \cite{Sutton:99}, the MAXQ framework \cite{Dietterich:00}, and the recursive composition of option models \cite{Silver:12}. 
Several approaches have been proposed to tackle the problem of learning good options. 
One approach is selecting frequently observed solution states as targets \cite{Stolle:02}. Another strategy involves employing gradient-based techniques to enhance the termination function for options \cite{Comanici:10, Mankowitz:16, Levy:12}. 
\cite{Levy:12} employed policy gradient optimization to learn a unified policy comprising intra-option policies, option termination conditions, and an option selection policy. 
Parametrized options can be learned by treating the termination functions as hidden variables and applying expectation maximization \cite{Daniel:16}. Intrinsic rewards have also been utilized to learn policies within options, while extrinsic rewards are used to learn the policy over options \cite{Kulkarni:16}. \cite{Bacon:17} proposed a method that jointly learns options and their associated policies using the policy gradient theorem. Additionally, a manager module operating on a slow time scale has been introduced to learn sub-goals, which are subsequently achieved by a worker module operating on a fast time scale \cite{Vezhnevets:17}.

% write about lihong li's paper
% write about different ways abstraction is done in literature
% 

%Bisimulation \citep{Givan:03, Ferns:12, Castro:19, Zhang:20} tries to aggregate states with similar behaviors. 

\section{Contrastive Learning of Sequential Proximity}
\label{app:method_contrastive}

\subsection{Hyperparameters for Constrastive Learning}

In the maze experiments, a four-layer fully connected network was employed, comprising of 256, 128, 64, and 32 neurons in each respective layer. The network utilized ReLU activations. The $\tau$ value was set to 30 for \ref{eq:lossNCE}. A batch size of 4096 was used, and the training process consisted of 1000 epochs. The learning rate was set to 1e-3, and a weight decay of 1e-5 was applied.

In the CIFAR environment experiments, a network architecture consisting of three CNN layers followed by three fully connected layers was utilized. The CNN layers had output channels of 64, 32, and 32, with kernel sizes of 4, 4, and 3 respectively. A stride and padding value of 2 was applied throughout. The fully connected layers had sizes of 256, 64, and 8 neurons respectively, employing ReLU activations. The $\tau$ value was set to 30 for \ref{eq:lossNCE}. A batch size of 4096 was used, and the training process spanned 8000 epochs. The learning rate was annealed using cosine annealing, starting from 1e-3 and gradually decreasing to a minimum value of 1e-6. A weight decay of 1e-5 was applied.

In the MiniGrid experiments, a network architecture comprising of three CNN layers followed by two fully connected layers was employed. The CNN layers had output channels of 64 and 32, with kernel sizes of 4, 4, and 2 respectively. Stride and padding values of 2 were uniformly used. The fully connected layers had sizes of 64 and 32 neurons respectively, employing ReLU activations. Following these layers, an LSTM layer and another fully connected layer were added, both with an output size of 32. The $\tau$ value was set to 30 for \ref{eq:lossNCE}. A batch size of 4096 was used, and the training process spanned 8000 epochs. The learning rate was annealed using cosine annealing, starting from 1e-3 and decreasing gradually to a minimum value of 1e-6. Additionally, a weight decay of 1e-5 was applied.

In the Minecraft experiments, the network architecture involves processing a sequence of the 32 most recent frames as input. The first stage of the network consists of four batch-normalized convolution layers with ReLU activation functions. These layers are structured as follows: Conv-Layer-1 with 16 feature maps, a kernel size of 4, a stride of 2, and zero padding of 1; Conv-Layer-2 with 32 feature maps, a kernel size of 4, a stride of 2, and zero padding of 1; Conv-Layer-3 with 64 feature maps, a kernel size of 3, and a stride of 2; and Conv-Layer-4 with 32 feature maps, a kernel size of 3, and a stride of 2. The resulting flattened latent representation (dimension: R32×288) from the convolution stage is then fed into an LSTM layer with 256 units for further processing.
The inverse tau value is set to 30 for the contrastive loss. A batch size of 1024 is used, and the training process involves 300 epochs. The learning rate is annealed using cosine annealing, starting from 1e-3 and gradually decreasing to a minimum value of 1e-6. Additionally, a weight decay of 1e-5 is applied.

\subsection{Hyperparameters for Sampling Strategies}
For the Gaussian sampling strategy, standard deviation ($\sigma$) of 15 is used for all experiments. For exponential distribution, we use a $\gamma$ of 0.99 for all experiments.

\section{Abstraction using Modern Hopfield Networks}
\label{app:learned_beta}

% TODO copied from CLOOB
\subsection{Review of Modern Hopfield Networks}
We briefly review  
continuous MHNs
that are used for deep learning architectures.
MHNs \citep{Ramsauer:21} map states to fixed points that 
correspond to abstract states.
Hopfield networks, introduced in the 1980s, are binary associative memories 
that played a significant role in popularizing artificial neural networks \citep{Amari:72,Hopfield:82,Hopfield:84}. 
These networks were designed to store and retrieve samples~\citep{Chen:86,Psaltis:86,Baldi:87,Gardner:87,Abbott:87,Horn:88,Caputo:02,Krotov:16}.
Since they exhibit continuity and differentiability, these functions are compatible with gradient descent in deep architectures.
As they can be updated in a single step, they can be activated similarly to other layers in deep learning.
With their exponential storage capacity, they are capable of handling large-scale problems effectively.
% , 
% and their storage capacity could be significantly enhanced by incorporating polynomial terms into the energy function \citep{Chen:86,Psaltis:86,Baldi:87,Gardner:87,Abbott:87,Horn:88,Caputo:02,Krotov:16}. 
In contrast to traditional binary memory networks, 
we employ continuous associative memory networks with exceptionally high storage capacity. 
These MHNs, utilized in deep learning architectures, 
have an energy function with continuous states and 
possess the ability to retrieve samples with just a single update \citep{Ramsauer:21}. 
The update rule of MHNs is guaranteed to converge to a fixed point. 
These fixed points correspond to attractors in the network's energy landscape, and the network dynamics converge towards these attractors when starting at an input pattern.
MHNs have already been demonstrated to be successful 
in diverse fields~\cite{Widrich:20, Seidl:22, Widrich:21, Paischer:22, Fuerst:21}.

We assume a set of patterns $\{\Bu_1,\ldots,\Bu_N\} \subset \dR^d$
that are stacked as columns to 
the matrix $\BU = \left( \Bu_1,\ldots,\Bu_N \right)$ and a 
state pattern (query) $\Bxi \in \dR^d$ that represents the current state. 
The largest norm of a stored pattern is
$M = \max_{i} \NRM{\Bu_i}$.
Continuous MHNs with state $\Bxi$
have the energy
\begin{align}
\rE  \ &=   \ -  \ \beta^{-1} \ \log \left( \sum_{i=1}^N
\exp(\beta \Bu_i^T \Bxi) \right)   \ +  \ \beta^{-1} \log N  \ + \  
\frac{1}{2} \ \Bxi^T \Bxi  \ +  \ \frac{1}{2} \ M^2 \ .
\end{align}
For energy $\rE$ and state $\Bxi$, the update rule 
\begin{align}
\label{eq:Amain_iterate}
\Bxi^{\nn} \ &= \ f(\Bxi;\BU,\beta) \ = \ \BU \ \Bp \ = \   \BU \ \soft ( \beta \BU^T \Bxi)
\end{align}
has been proven to converge globally  
to stationary points of the energy $\rE$, 
which are almost always local minima 
\citep{Ramsauer:20,Ramsauer:21}.
The update rule Eq.~\eqref{eq:Amain_iterate}
is also the formula of the well-known transformer attention mechanism
\citep{Ramsauer:20,Ramsauer:21}, therefore Hopfield retrieval and
transformer attention coincide.

The {\em separation} $\Delta_i$  of a 
pattern $\Bu_i$ is defined as its minimal dot product difference to any of the other 
patterns:
$\Delta_i = \min_{j,j \not= i} \left( \Bu_i^T \Bu_i - \Bu_i^T \Bu_j \right)$. 
A pattern is {\em well-separated} from the data if $
 \Delta_i  \geq \frac{2}{\beta N} + \frac{1}{\beta} \log \left( 2 (N-1)  N  \beta  M^2 \right)$.

In the case where the patterns $\Bu_i$ exhibit good separation, the iteration defined by Eq.~\eqref{eq:Amain_iterate} converges to a stable point that is in close proximity to one of the stored patterns. However, if certain patterns are similar to each other and lack clear separation, the update rule Eq.~\eqref{eq:Amain_iterate} converges to a stable point that is close to the mean of those similar patterns. This particular fixed point represents a metastable state of the energy function, effectively averaging over the similar patterns.

According to the next theorem, when the patterns are well separated, the update rule defined by Eq.~\eqref{eq:Amain_iterate} typically achieves convergence within a single update. Additionally, the theorem asserts that the retrieval error is exponentially small with respect to the separation $\Delta_i$ between the patterns.

\begin{theoremA}[Modern Hopfield Networks: Retrieval with One Update \citep{Ramsauer:21}]
\label{th:AoneUpdate}
With query $\Bxi$, after one update the distance of the new point $f(\Bxi)$
to the fixed point $\Bu_i^*$ is exponentially small in the separation $\Delta_i$.
The precise bounds using the Jacobian $\rJ = \frac{\partial
  f(\Bxi)}{\partial \Bxi}$ and its value $\rJ^m$ in the mean value
theorem are:
\begin{align}
  &\NRM{f(\Bxi) \ - \ \Bu_i^*}
  \ \leq \  \NRM{\rJ^m}_2 \ \NRM{\Bxi \ - \ \Bu_i^*}  \ , \\
  &\NRM{\rJ^m}_2  \ \leq \
  2 \ \beta \ N \ M^2 \ (N-1) \exp(- \ \beta \
  (\Delta_i \ - \ 2 \  \max \{ \NRM{\Bxi  \ - \ \Bu_i} , \NRM{\Bu_i^* \ - \ \Bu_i} \}  \ M) )\ .
\end{align}
For given $\epsilon$ and 
sufficient large $\Delta_i$, we have $\NRM{f(\Bxi) \ - \ \Bu_i^*} < \epsilon$,
that is, retrieval with one update.
The retrieval error $\NRM{f(\Bxi) \ - \ \Bu_i}$ of pattern $\Bu_i$
is bounded by
\begin{align}
  \NRM{f(\Bxi) \ - \ \Bu_i} \ &\leq \ 2 \ (N-1) \ \exp(- \ \beta \ 
  (\Delta_i \ - \ 2 \   \max \{ \NRM{\Bxi  \ - \ \Bu_i} , \NRM{\Bu_i^* \ - \ \Bu_i} \} 
  \ M) )  \ M  \ .
 \end{align}
\end{theoremA}
For a proof see \citep{Ramsauer:20,Ramsauer:21}.

Our objective is to store a potentially extensive collection of embeddings. To accomplish this, we initially establish the concept of storing and retrieving patterns within a contemporary Hopfield network.

\begin{definitionA}[Pattern Stored and Retrieved \citep{Ramsauer:21}]
We assume that around every pattern $\Bu_i$ a sphere $\rS_i$ is given.
We say $\Bu_i$ {\em is stored} if there is a single fixed point $\Bu_i^* \in \rS_i$ to
which all points $\Bxi \in \rS_i$ converge,
and  $\rS_i \cap \rS_j = \emptyset$ for $i \not= j$.
We say $\Bu_i$ {\em is retrieved} for a given $\epsilon$ if 
iteration (update rule) Eq.~\eqref{eq:Amain_iterate} gives
a point $\tilde{\Bx}_i$ that is at least 
$\epsilon$-close to the single fixed point $\Bu_i^* \in \rS_i$. 
The retrieval error is $\NRM{\tilde{\Bx}_i - \Bu_i}$.
\end{definitionA}

Similar to classical Hopfield networks, we focus on patterns that lie on a sphere, meaning patterns with a constant norm. In the case of randomly selected patterns, the number of patterns that can be stored increases exponentially with the dimensionality $d$ of the pattern space ($\Bu_i \in \mathbb{R}^d$).

\begin{theoremA}[Modern Hopfield Networks: Exponential Storage Capacity \citep{Ramsauer:21}]
\label{th:Astorage}
We assume a failure probability $0<p\leq 1$ and randomly chosen patterns 
on the sphere with radius $M:=K \sqrt{d-1}$. 
We define $a := \frac{2}{d-1}  (1 + \ln(2 \beta K^2 p (d-1)))$, 
$b := \frac{2  K^2  \beta}{5}$,
and $c:= \frac{b}{W_0(\exp(a + \ln(b))}$,
where $W_0$ is the upper branch of the Lambert $W$ function 4.13,
and ensure $c \geq \left( \frac{2}{ \sqrt{p}}\right)^{\frac{4}{d-1}}$.
Then with probability $1-p$, the number of random patterns 
that can be stored is  
\begin{align} 
 \label{eq:ACapacityM}
    N \ &\geq \ \sqrt{p} \ c^{\frac{d-1}{4}}  \ .
\end{align}
Therefore it is proven for $c\geq 3.1546$ with
$\beta=1$, $K=3$, $d= 20$ and $p=0.001$ ($a + \ln(b)>1.27$)
and proven for $c\geq 1.3718$ with $\beta = 1$, $K=1$, $d = 75$, and $p=0.001$
($a + \ln(b)<-0.94$).
\end{theoremA}
For a proof see \citep{Ramsauer:20,Ramsauer:21}.

The validity of employing continuous modern Hopfield networks in substituting retrieved embeddings instead of the original embeddings for large batch sizes is supported by this theorem. Even with a substantial number of embeddings, reaching into the hundreds of thousands, the continuous modern Hopfield network demonstrates its capability to retrieve the embeddings, provided that the dimension of the embeddings is sufficiently large.

\subsection{Hyperparameters for Hopfield Network}
For the fixed $\beta$ visualisation experiments Fig~\ref{fig:increasing_beta} on Maze2D environment, we use $\beta$ values of 20, 25 and 35. 

For the fixed $\beta$ visualisation experiments, Fig~\ref{fig:increasing_beta} on RedBlueDoor environment, we use $\beta$ values of 100, 150 and 200 (left to right in Fig ~\ref{fig:increasing_beta}. 

For the fixed $\beta$ visualisation experiments, Fig~\ref{fig:increasing_beta} on Minecraft environment, we use $\beta$ values of 30, 35 and 50.

We currently do not have any learnable weights in the hopfield, but a hopfield network can be trained with learnable weights also. 

\subsection{Learned Beta Training}
For all our experiments we use a single fully connected network as the $\beta$ network. This $\beta$ network has a sigmoid output neuron, which predicts a value between 0 and 1. Further, we multiply this output with $\beta_{max}$, which is the maximum value the $\beta$ can have. Thus, bounding the $\beta$ value. For all our experiments we use the $\beta_{max}$ value of 200. 

We use a single fully connected network after hopfield network. 
In the experiment, the learning rate was set to 1e-3. A weight decay of 1e-5 was applied to control the regularization of the model's weights. The training process spanned a total of 1000 epochs, with each epoch representing a complete iteration through the dataset. Furthermore, a masking ratio of 0.3 was employed, indicating the proportion of the input or data that was randomly masked or hidden during training.

\section{Policy and Planning Experiments}
\label{app:experiments}

\subsection{Environment Details}
\paragraph{CifarEnv} is an environment consisting of images and classes from the CIFAR-100 \cite{Krizhevsky:09} dataset. 
This environment is designed to verify the abstract representation as the true number of abstract states and their relation is known.
We select 10 classes and define a transition structure for various goals and tasks (see Figure \ref{fig:cifar_planning}(a)).
For example, in \textit{GoSchool} the agent starts in state \textit{bed} and receives reward if it reaches the state \textit{school}. Other tasks include, \textit{GoMarket, GoOffice, GoHouse}. 
The agent stays in the same state for $n$ timesteps (where $n=8$ for our experiments) in this environment, 
and is presented with an image sampled from the respective class in each timestep.
After $n$ timesteps the agent then selects an action in order to transition to the next state.
For example, if the agent is in the state \textit{house} then $n$ different images of this class are sampled from CIFAR-100 sequentially. 
We randomly sample $100,000$ timesteps from this environment and build a dataset for constrastive abstraction learning. 

\paragraph{Maze2D-large} is an environment from Mujoco Suite \cite{Todorov:2012:mujoco}. 
A robot must navigate to a given goal state from a random starting position.
The state is an 4-dimensional vector and the action space is continuous, controlling the agents directional movement. 
We use trajectories from the D4RL dataset \cite{Fu:2020:d4rl} for our experiments.

\paragraph{RedBlueDoor} is part of the Minigrid library \cite{Chevalier:2018:minigrid} and is a partially observable environment.
The agent has access to a partial view of the environment and can select from $7$ discrete actions.
We generate a dataset by storing all interactions during training of a policy with Proximal Policy Optimization (PPO) \cite{Schulman:17}.

\paragraph{Minecraft} is an environment based on the popular sandbox game Minecraft.
The environment was released together with a dataset of demonstrations from human players performing various tasks for the MineRL competition \cite{Guss:19comp}.
Several tasks, such as mining a diamond, were defined for the competition.
We use the environment and the dataset of human demonstrations released for the MineRL competition \cite{Guss:19comp}.

\subsection{Policy experiments}
We use the same PPO hyperparameters across all the methods. The clip coefficient was set to 0.2, which limits the size of the policy update to prevent drastic changes. The entropy coefficient was set to 0.0001, encouraging exploration by adding an entropy term to the objective function. The learning rate was set to 1e-5, indicating the step size used in the optimization process. There was no annealing applied to the learning rate or clipping parameter throughout the training process. Additionally, to mitigate the impact of large gradients, the gradient was clipped if its norm exceeded 0.5. The experiments were averaged over 10 different seeds. 

\subsection{Learning Policy between fixed points}
We learn policy between fixed points using behavior cloning. We first map each state to the corresponding fixed point. Further, we pair the state with the next fixed point in the sequence, thus obtaining a dataset where each sample has the following information, 
($s_t, \Bxi_{t}, \Bxi_{t'}, a_t $). 
Further, we condition the policy network with the current state $s_{t}$ 
and future fixed point $\Bxi_{t'}$ and predict the action $a_t$ in the dataset. 

These policy between fixed points are used for both MetaPolicy and Planning experiments. 

\subsection{Planning}
For the CIFAR experiments, we use a threshold of 0.68 to prune the connections in the graph. For Maze2D experiments, we use a threshold of 0.6. We also remove self-connection by removing most similar connections to the current node using another threshold (0.97, 0.9 respectively). Instead of using a threshold, we can also use the dataset to prune connections. We can look for consecutive fixed points in the entire dataset and remove connections which were not found.

\section{Compute and Software Libraries}

\subsection{Compute Details}
For all experiments, during development 2 to 3 nodes each with 4 GPUs of an 
internal GPU cluster were used for roughly 
six months of GPU compute time (Nvidia A100). 

\subsection{Software Libraries}
We are thankful towards the developers of  Stable-Baselines \cite{Raffin:21}, PyTorch \cite{pytorch2019}, OpenAI Gym \cite{Brockman:16}, Numpy \cite{harris2020array}, Matplotlib \cite{Hunter:2007} and Minecraft \cite{Guss:19}. We also used code from \cite{patil:22}.

% \bibliography{minimo}
% \bibliographystyle{plainnat}

\end{document}